\documentclass[runningheads]{llncs}
\usepackage{graphicx}
\usepackage{amssymb}
\usepackage{epstopdf}
\usepackage{booktabs}
\usepackage{subfigure}
\usepackage{cite}
\begin{document}
\title{A Hierarchically Feature Reconstructed Autoencoder for Unsupervised Anomaly Detection}
%
%
\author{Honghui Chen\inst{1} \and
Pingping Chen\inst{1,*} \and
Huan Mao\inst{1} \and
Mengxi Jiang\inst{2}}
%


\authorrunning{H. Chen et al.}
%
\institute{Department of Physics and Information Engineering, Fuzhou University, Fuzhou 350108, China
\and
School of Advanced Manufacturing, Fuzhou University, Jinjiang, 362251, China
*Corresponding author:\email{ppchen@fzu.edu.cn}\\ }

%
%
%
\maketitle              
\begin{abstract}
Anomaly detection and localization without any manual annotations and prior knowledge is a challenging task under the setting of unsupervised learning. The existing works achieve excellent performance in the anomaly  detection,  but   with complex networks or cumbersome pipelines.
To address this issue, this paper explores a simple but effective architecture in the anomaly detection. It consists of a well pre-trained encoder to extract hierarchical feature representations and a decoder to reconstruct these intermediate features from the encoder. In particular, it does not require any data augmentations and anomalous images for training. The anomalies can be detected when the decoder fails to reconstruct features well, and then errors of hierarchical feature reconstruction are aggregated into an anomaly map to achieve anomaly localization. The difference comparison between  those features of encoder and decode lead  to more accurate and robust localization results than the comparison in single feature or pixel-by-pixel comparison in the conventional works. Experiment results show that the proposed method outperforms the state-of-the-art methods on MNIST, Fashion-MNIST, CIFAR-10, and MVTec Anomaly Detection datasets on both anomaly detection and localization.
\keywords{Deep Learning  \and Anomaly Detection \and Unsupervised Learning \and Feature Reconstruction.}
\end{abstract}
\section{Introduction}

Anomaly detection is crucial in computer vision, especially in industrial inspection \cite{bergmann2019mvtec}. Detecting defects and locating them ensures product quality. However, abnormal data in real scenarios are difficult to know in advance due to scarcity, unknown, and randomness. Hence, unsupervised anomaly detection and localization are the current major research areas. Since only anomaly-free data are available \cite{chandola2009anomaly}, the approach of image-level binary classification is inefficient \cite{chen2001one, scholkopf2002learning}. Generative Adversarial Networks (GAN) \cite{akcay2018ganomaly, schlegl2019f} and Autoencoder (AE) \cite{abati2019latent, schlegl2019f, bergmann2018improving, vasilev2020q, venkataramanan2019attention} are commonly used for anomaly detection. However, the per-pixel reconstruction error function used by these methods fails in locating subtle anomalies due to inaccurate reconstruction \cite{bergmann2018improving}. Recent studies have achieved better performance by utilizing well-pre-trained models on large datasets \cite{venkataramanan2019attention, burlina2019s, perera2019deep, cohen2020sub, rippel2021modeling}. To overcome the limitation of training models from scratch, teacher-student networks have been utilized with some success, but suffer from limited localization performance and high training cost \cite{bergmann2020uninformed}. Recent works based on teacher-student networks directly use information from multiple intermediate layers to achieve better performance and reduce training costs \cite{wang2021student}. However, features are not completely utilized as soft targets to train, as there are still losses between student and teacher in the training phase.
Previous studies have achieved state-of-the-art performance in anomaly detection, but their complex networks and cumbersome detection processes make them less than ideal \cite{defard2021padim,wang2021student}. To address this issue, we propose a streamlined approach that leverages the similarities in the distribution of anomaly-free images. We can effectively detect and locate anomalies without requiring manual annotations by using autoencoders or teacher-student networks. Compared to learning feature representation from the teacher network, the decoder in the autoencoder fully utilizes feature information by directly decoding features.

\begin{figure}[!t]
    \vspace {-4mm}
	\centering
		\includegraphics[width=5.9cm]{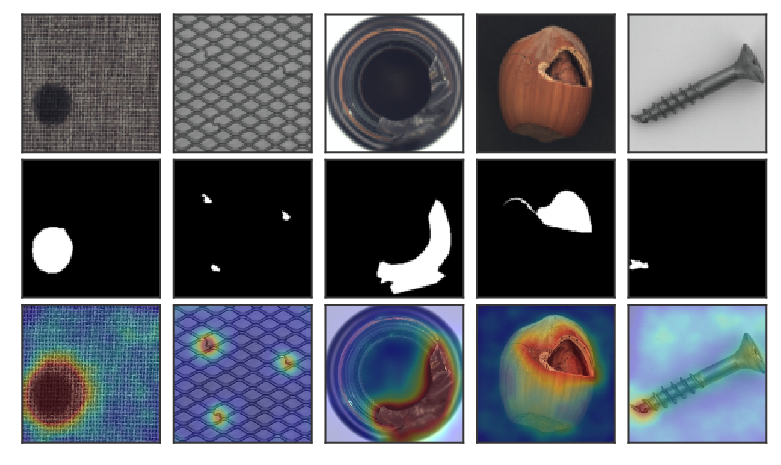}
	\caption{The visualization results of our method on the MVTecAD dataset. Each column represents a category in the dataset, and rows from top to bottom correspond to defective images, ground truth regions, and anomaly heat maps inferred by our method.}
	\label{FIG:1}
\end{figure}

Inspired by \cite{defard2021padim,wang2021student,salehi2021multiresolution,rippel2021modeling,cohen2020sub}, we improve the structure of the encoder-decoder and introduce  hierarchical feature reconstruction for unsupervised anomaly detection and localization.
Specifically, we first use a deep neural network pre-trained on ImageNet\cite{krizhevsky2012imagenet} as an encoder, which can extract hierarchical feature representations of the image in the latent space. Then, we adopt a decoder to obtain the output features of the encoder to reconstruct these features instead of an anomaly-free image at the pixel level.
For feature reconstruction, we align the size of each corresponding feature in the encoded features and reconstructed features separately. We finally calculate the loss based on their differences to train the decoder to learn the feature distribution of anomaly-free images.
Fig. \ref{FIG:1} shows the visualized results of the proposed method on the MVTec Anomaly Detection (MVTecAD) dataset \cite{bergmann2019mvtec}. We compare our method with the state-of-the-art methods on MNIST \cite{lecun1998gradient}, Fashion-MNIST \cite{xiao2017fashion}, CIFAR-10 \cite{krizhevsky2009learning} and MVTecAD datasets. The results show that  our method achieves   competitive results that exceeds state-of-the-art methods on the average AUROC of localization and detection.

The main contributions of this paper are summarized as follows:
(i) We propose an effective encoder-decoder framework without complex modules, which leverages a pre-trained feature extractor and hierarchical feature reconstruction errors to compute the anomaly map.
(ii) We propose to constrain the intermediate features of the proposed autoencoder in the training phase to learn embeddings of anomaly-free images in different feature spaces.
(iii) The proposed method  outperforms the state-of-the-art methods on the MNIST, Fashion-MNIST, CIFAR-10, and MVTec Anomaly Detection dataset in both anomaly detection and localization.

\section{Related work}\label{sec:related work}

\subsection{Methods Based on Reconstruction}

In reconstruction-based methods, autoencoders and GANs were   widely applied in anomaly detection on an assumption that a network trained only on normal data cannot reconstruct abnormal data well. Autoencoders train an encoder to learn the feature representation of the normal image, and a decoder to reconstruct normal images. When inputting an abnormal image, anomalies will be localized by comparing the pixel residual between the reconstructed image and the input image. However, the simple pixel-wise comparison in the autoencoder will cause imperfect reconstruction effects and be easily disturbed by image noise. Bergmann et al.\cite{bergmann2018improving} proposed a perceptual loss function based on structural similarity instead of simply comparing single pixel values to train an autoencoder, which achieves significant performance. Shashanka et al.\cite{venkataramanan2019attention} proposed a convolutional adversarial variational autoencoder with guided attention (CAVGA), which processes anomaly detection and localization with attention maps.

GAN methods train a generator to learn the distribution of normal images in an adversarial training process\cite{goodfellow2014generative}. In the inference phase, the discriminator determines whether the input image is normal. AnoGAN\cite{schlegl2017unsupervised} was the first research to use the GAN for image anomaly detection, trying to find a feature vector in the latent space to make the generated image most similar to the input image. GANomaly\cite{akcay2018ganomaly} and f-anoGAN\cite{schlegl2019f} attempted to add an additional encoder in addition to the autoencoder to reduce the inference time, and both add feature information as a loss function during the training phase.

\subsection{Methods Based on Knowledge Distillation}

The distribution of training features \cite{bergmann2020uninformed} was implicitly modeled  with a teacher-student approach, where   an ensemble of student networks is trained to mimic the teacher's output and computing anomaly maps based on the predictive variance and regression error of the student networks. However, its localization performance is limited by the fixed size of the patch and the complex network   suffers from expensive training cost. Some latest methods \cite{salehi2021multiresolution, wang2021student} attempted to distill the intermediate feature  at various layers from the well pre-trained network and directly computed the anomaly map. This process greatly simplifies the training and reduces the inference time via only two forward propagation.

\section{Methodology}

\subsection{Overview}

In this section, we mainly introduce the framework of the proposed autoencoder based on the hierarchical feature reconstruction.
We propose a simple yet effective method for unsupervised anomaly detection, with an encoder-decoder architecture as our basic paradigm, where the encoder is loaded with a well pre-trained model and a hierarchical feature reconstruction is introduced to train the decoder and compute the anomaly map.

\subsubsection{Unsupervised Anomaly Detection}
As mentioned in Section. \ref{sec:related work}, most unsupervised anomaly detection methods can be roughly divided into three categories, i.e., encoder-decoder, teacher-student network, and memory bank. Generally, the autoencoder expects to reconstruct the input image from the feature vector extracted by the encoder, which requires only one forward propagation in the inference phase.
The teacher-student network aims to supervise the learning of the student network using the feature vectors output from the teacher network, but the forward propagation of  both the networks  needs to be completed separately in the inference phase.
The memory bank approach collects features from training images and requires high computational overhead in large training sets.

Among them, the encoder-decoder architecture of the autoencoder is used as   the basic paradigm of our approach  for its simplicity.
Given an input image $x\in {\mathbb{R}^{c\times{w}\times{h}}}$, where $c$, $w$ and $h$ denote the numbers of channels, width and height of the input image, respectively.
 A basic autoencoder consists of an encoder to extract a feature representation $z$ and a decoder to generate the reconstruction of the input image $\hat{x}\in {\mathbb{R}^{c\times{w}\times{h}}}$. The process of network forward propagation is formulated as:
\begin{small}
\begin{equation}
\hat{x} = D(E(x)) = D(z),
\end{equation}
\end{small}
where $E(\cdot)$ and $D(\cdot)$ denote the encoder function and the decoder function, respectively. During evaluation, anomalies are detected by computing a residual map through the per-pixel comparison between $x$ and $\hat{x}$. However, the pixel-level reconstruction is prone to inaccurate reconstruction, while feature representation in the feature space of the image is more robust\cite{bergmann2018improving,akcay2018ganomaly,schlegl2019f}. Thus, we modify the structure of the autoencoder and introduce  hierarchical feature reconstruction to capture different kinds of anomalies.

\begin{figure*}[!t]
    \vspace {-4mm}
	\centering
		\includegraphics[width=\textwidth]{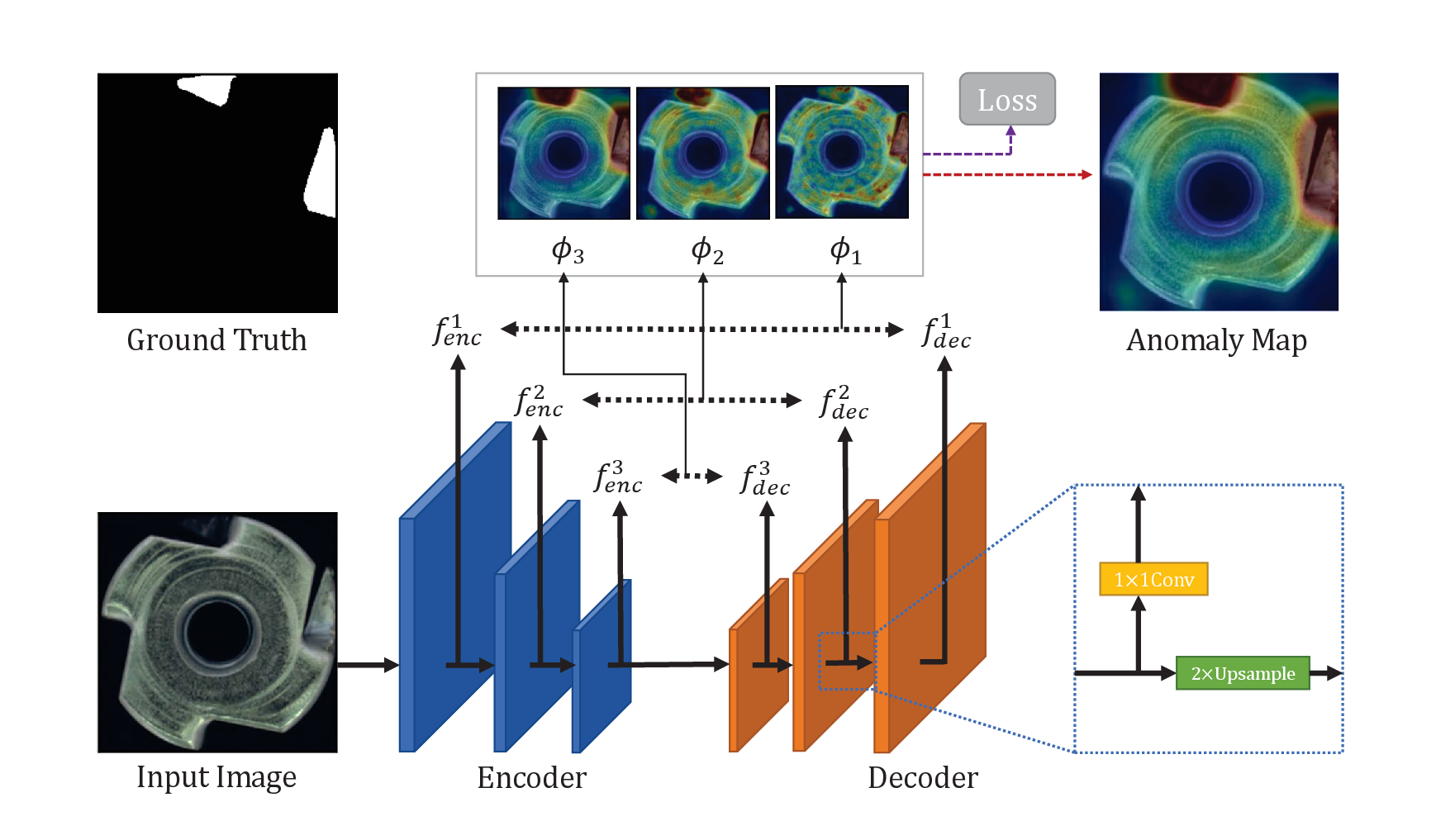}
	\caption{ An overview of our proposed framework. The encoder extracts the feature representations of different layers of the input image, and the decoder decodes the high-dimensional output feature of the encoder to reconstruct the features of different layers. The residual feature maps $\{ \phi_1, \phi_2, \phi_3\}$ between the hierarchical features of the encoder and the decoder is expressed as a hierarchical feature reconstruction loss in the training phase to guide the decoder to reconstruct the features of the normal image, and in the inference phase as an anomaly map to detect and locate anomalies. }
\label{FIG:2}
\end{figure*}
\subsubsection{Hierarchical Features Reconstruction}

In the anomaly detection, the types of anomalies are unknown and diverse, and distributions of these anomalies are inconsistent. Thus, it is necessary to distinguish them in different feature spaces.
The teacher-student network can  supervise the student network and learn intermediate features \cite{salehi2021multiresolution, wang2021student}. Moreover, in this paper, we expect the decoder to perfectly reconstruct the features from the encoder of the corresponding layer in the autoencoder framework.

As shown in Fig. \ref{FIG:2}, we use the first three layers of ResNet\cite{he2016deep} as the encoder which is responsible for extracting a group of hierarchical feature representations $F_{enc}=\{f_{enc}^1,f_{enc}^2,\cdots,f_{enc}^K\}$,  $f_{enc}^k\in {\mathbb{R}^{c_k\times{w_k}\times{h_k}}}$ is the $k$-th encoded feature map, $k\in \{1, 2, \cdots, K\}$ and the number of extracted feature maps $K$ is set to 3 in our model.
It is worth mentioning that a deeply pre-trained network can lead to better performance in anomaly detection  compared to training a feature extractor from scratch\cite{cohen2020sub,rippel2021modeling}. Therefore, our encoder loads the pre-trained model on a large dataset and freezes the parameters from training.
The feedforward process of the encoder can be formulated as:
\begin{small}
\begin{equation}
F_{enc}=\{f_{enc}^1,f_{enc}^2,\cdots,f_{enc}^K\}=E(x).
\end{equation}
\end{small}
We then design the decoder to reconstruct hierarchical features that correspond to multi-level encoded features.
The group of reconstructed features is denoted as $F_{dec}=\{f_{dec}^1,f_{dec}^2,\cdots,f_{dec}^K\}$, where $f_{dec}^k\in {\mathbb{R}^{c_k\times{w_k}\times{h_k}}}$ is the reconstructed feature of the $k$-th encoded feature, $k\in \{1, 2, \cdots, K\}$. Specifically, we use a group of convolution blocks with the same kernel size but different channels at each level and resize the feature using an upsampling layer.
To integrate the channel information of features, an extra convolution layer with a kernel size of $1 \times 1$ is applied before each reconstructed feature $f_{dec}^k$ outputs. The procedure of the $k$-th reconstructed feature can be summarized as:
\begin{small}
\begin{equation}
f_{dec}^k=Conv_{1 \times 1}(Convs(\tilde{f}_{dec}^{k-1})),
\end{equation}
\end{small}
and
\begin{small}
\begin{equation}
\tilde{f}_{dec}^{k-1}=Upsampling(f_{dec}^{k-1}),
\end{equation}
\end{small}
where $Convs(\cdot)$ denotes a group of convolution blocks, $Conv_{1 \times 1}(\cdot)$ is a $1 \times 1$ convolution, $\tilde{f}_{dec}^{k-1}$ denotes the $k$-$1$-th reconstructed feature sampled up to the size of the $k$-$1$-th encoded feature and $Upsampling(\cdot)$ is a nearest neighbor upsampling layer. In addition, $f_{enc}^K$ is fed directly into the decoder without the upsampling operation.

\subsubsection{Calculation of Residual Maps}

Then, the hierarchical differences between the encoded features and the reconstruction features are aggregated to compute an anomaly map.
Specifically, we measure the element-wise difference between the encoded feature $f_{enc}^k$ and the reconstructed feature $f_{dec}^k$ after L2 normalization, which allows us to obtain the residual map  and then locate anomalies.
The calculation of the residual map $\phi_k$ is formulated as:
\begin{small}
\begin{equation}
\phi_k = \frac{1}{2}\Vert \hat{f}_{enc}^k - \hat{f}_{dec}^k \Vert_2^2,
\end{equation}
\end{small}
and
\begin{small}
\begin{equation}
\hat{f}_{enc}^k = \frac{f_{enc}^k}{\Vert f_{enc}^k \Vert_2}, \qquad \hat{f}_{dec}^k = \frac{f_{dec}^k}{\Vert f_{dec}^k \Vert_2},
\end{equation}
\end{small}
where $\hat{f}_{enc}^k$ and $\hat{f}_{dec}^k$ are the normalized encoded and reconstructed features, respectively,
$\Vert \cdot \Vert_2$ denotes the L2 distance and $\phi_k\in {\mathbb{R}^{c_k\times{w_k}\times{h_k}}}$ denotes the $k$-th residual map.
Fig. \ref{FIG:2} shows the visualized qualitative results of the residual maps $\{ \phi_1, \phi_2, \phi_3\}$ of an anomalous image. The residual map $\phi_1$ on bottom-level features generates larger values on some shallow features such as edges, and inevitably produces some values on natural noise regions. On the contrary, the   map $\phi_3$ of top-level features has a smoother   variation of the value and a larger receptive field with rich semantic information.

In the training phase with only anomaly-free images, the loss function is constructed according to the maps $\{ \phi_1, \phi_2, \phi_3\}$ at different levels to make their values approach zero. In the inference phase, the anomaly map is obtained by integrating these residual maps to locate anomalies. Note that due to this concise network architecture, we only need one forward propagation to achieve anomaly detection and localization. For the architecture of the decoder, we set up a group of ablation experiments to explore the most effective decoder architecture.

\subsection{Training Phase}

In the entire network architecture, the parameters of the encoder are frozen and do not participate in the training. What we need to train is the feature reconstruction ability of the decoder. Given a training dataset $D_{train}=\{x_1,x_2,\cdots,x_n\}$,   the decoder aims to  reconstruct the features of the normal image, but cannot correctly reconstruct the features of the abnormal one.
The multi-level residual maps $\{ \phi_1, \phi_2, \phi_3\}$ provide a good training constraint to measure the reconstruction quality of the decoder.
We denote the value on the point $(i,j)$ of   $\phi_k$ as $d_k(i,j)$, where $i\in { \{1,\cdots,w_k \} }$, $j\in { \{1,\cdots,h_k \} }$, and the $k$-th feature loss $L_k$ is expressed as the average value of $\phi_k$ under its feature size $w_k \times h_k$, given by:
\begin{small}
\begin{equation}
L_k = \frac{1}{w_k \times h_k}\displaystyle\sum_{n=1}^{c_k}\displaystyle\sum_{j=1}^{h_k}\displaystyle\sum_{i=1}^{w_k}d_k(i,j),
\end{equation}
\end{small}
where $c_k$, $w_k$ and $h_k$ are the number of channel, width and height of the $k$-th feature map, respectively. Then, the total loss $L$ is defined as the average of all feature losses:
\begin{small}
\begin{equation}
L = \frac{1}{K}\displaystyle\sum_{k=1}^{K}L_k.
\end{equation}
\end{small}

\subsection{Testing Phase}

After the training, the decoder  learns the distribution of normal data and then   compute an anomaly map to detect and locate anomalies.
Given a test image $J$, we resize each residual map $\phi_k$ between the encoded feature and the reconstructed feature to the size $w \times h$ of the input image by bilinear interpolation.
 The summation of all resized residual maps finally yields the anomaly map as:
\begin{small}
\begin{equation}
\phi_k^{'} = Resize(\frac{1}{w_k \times h_k}\displaystyle\sum_{n=1}^{c_k}\phi_k),
\end{equation}
\end{small}
and
\begin{small}
\begin{equation}
f_a = G_\sigma(\displaystyle\sum_{k=1}^{K}\phi_k^{'}),
\end{equation}
\end{small}
where $\phi_k^{'} \in {\mathbb{R}^{1 \times{w\times{h}}}}$ denotes the $k$-th resized residual map with the size of $w \times h$ and $f_a \in {\mathbb{R}^{1 \times{w\times{h}}}}$ denotes the final anomaly map with the same size of $\phi_k^{'}$. To suppress random noise during image input, we use a Gaussian filter $ G_\sigma $ with standard deviation of $\sigma$ on the summation of $\phi_k^{'}$ and set $\sigma$ to 4.

Because the anomaly map $f_a$ aggregated by multi-scale feature residuals contains both spatial and semantic information, the regions with larger values on the anomaly map $f_a$ can be considered as the anomaly regions in the input image. In practice, anomaly localization can be completed by selecting an appropriate threshold. For anomaly detection, the anomaly score and the anomaly map are related, i.e. the anomaly score should be larger when anomaly regions with larger values appear on the anomaly map. Here, the maximum value in the map $f_a$ is defined as the anomaly score:
\begin{small}
\begin{equation}
A = Max(f_a).
\end{equation}
\end{small}
\section{Experiments}
To demonstrate the effectiveness of the proposed approach, we evaluate our model on MVTecAD\cite{bergmann2019mvtec}, MNIST\cite{lecun1998gradient}, Fashion-MNIST\cite{xiao2017fashion} and CIFAR-10\cite{krizhevsky2009learning} datasets for anomaly detection, and on the MVTecAD dataset for anomaly localization.
We used SGD optimizer (momentum=0.9, initial learning rate=0.4, weight decay=0.0001) for network optimization\cite{sutskever2013importance}. Pre-trained ResNet from ImageNet is used as the encoder and kept its parameters fixed during training. Each layer contained three convolution blocks with kernel sizes of  $1 \times 1$, $3 \times 3$, and $1 \times 1$, which were followed by batch normalization and LeakyReLu activation function with a negative slope of 0.1. To maintain consistent feature size, we applied nearest-neighbor sampling with a scale factor of 2 between layers of different scales.

\subsection{Experimental Results}


\subsubsection{Anomaly Detection}


To classify whether an image is anomalous, classification datasets like MNIST, Fashion-MNIST, and CIFAR-10 can be used for anomaly detection tasks by setting the target class to a normal class and other classes as abnormal. Tabel. \ref{det_auroc_mvtec} shows the performance of anomaly detection results of our method and the existing state-of-the-art methods. Our method outperforms existing state-of-the-art methods in terms of anomaly detection on the MVTecAD dataset, with an improvement from 7\% to 28\% on average AUROC. The proposed method excels at correctly classifying anomalous and normal images in categories such as carpet, leather, bottle, hazelnut, and metal nut. As shown in Table. \ref{det_3datasets}, the performance of our method on three other classification datasets also appears competitive compared to the MKD method. Overall, our proposed method's excellent anomaly detection performance on four datasets demonstrates its effectiveness and reliability in exploiting the difference in feature reconstruction at different resolutions and dimensions in image-level anomaly detection.

\begin{table*}[]
\centering
\caption{AUROC \cite{bergmann2020uninformed} for anomaly detection on MVTecAD dataset.}
    \vspace {-2.5mm}
    \label{det_auroc_mvtec}
    \setlength{\tabcolsep}{4.5pt}
\newcommand{\tabincell}[2]{\begin{tabular}{@{}#1@{}}#2\end{tabular}}
\resizebox{0.75\linewidth}{!}{
\begin{tabular}{cccccccc}
\toprule
           & CAVGA-D$_u$\cite{venkataramanan2019attention} & VAE$_{grad}$\cite{dehaene2020iterative} & SPADE\cite{cohen2020sub} & \tabincell{c}{Patch\\SVDD\cite{yi2020patch}} & MKD\cite{salehi2021multiresolution}   & \tabincell{c}{Ours\\(R18)} & \tabincell{c}{Ours\\(wR50)} \\ \hline
           Carpet     & 0.73     & 0.67      & 0.928 & 0.929          & 0.793 & \textbf{1.000} & \textbf{1.000} \\
           Grid       & 0.75     & 0.83      & 0.473 & 0.946          & 0.781 & 0.976          & \textbf{0.988} \\
           Leather    & 0.71     & 0.71      & 0.954 & 0.909          & 0.951 & \textbf{1.000} & \textbf{1.000} \\
           Tile       & 0.70     & 0.81      & 0.965 & 0.978          & 0.916 & 0.979          & \textbf{0.986} \\
           Wood       & 0.85     & 0.89      & 0.958 & 0.965          & 0.943 & \textbf{0.996} & 0.986          \\
           Bottle     & 0.89     & 0.86      & 0.972 & 0.986          & 0.994 & \textbf{1.000} & \textbf{1.000} \\
           Cable      & 0.63     & 0.56      & 0.848 & 0.903          & 0.892 & 0.988          & \textbf{0.989} \\
           Capsule    & 0.83     & 0.86      & 0.897 & 0.767          & 0.805 & 0.932          & \textbf{0.936} \\
           Hazelnut   & 0.84     & 0.74      & 0.881 & 0.920          & 0.984 & \textbf{1.000} & \textbf{1.000} \\
           Metal nut  & 0.67     & 0.78      & 0.710 & 0.940          & 0.736 & \textbf{1.000} & 0.999          \\
           Pill       & 0.88     & 0.80      & 0.801 & 0.861          & 0.827 & \textbf{0.979} & 0.977          \\
           Screw      & 0.77     & 0.71      & 0.667 & 0.813          & 0.833 & 0.876          & \textbf{0.960} \\
           Toothbrush & 0.91     & 0.89      & 0.889 & \textbf{1.000} & 0.922 & 0.997          & 0.983          \\
           Transistor & 0.73     & 0.70      & 0.903 & 0.915          & 0.856 & 0.964          & \textbf{0.997} \\
           Zipper     & 0.87     & 0.67      & 0.966 & 0.979          & 0.932 & 0.960          & \textbf{0.986} \\ \hline
           Mean       & 0.78     & 0.77      & 0.855 & 0.921          & 0.877 & 0.976          & \textbf{0.986} \\
\bottomrule
\end{tabular}
}
\end{table*}

\begin{table}[!h]
\centering
\newcommand{\tabincell}[2]{\begin{tabular}{@{}#1@{}}#2\end{tabular}}
\caption{The anomaly detection results comparing with the state-of-the-art methods on MNIST, Fashion-MNIST, and CIFAR-10 datasets}
    \vspace {-2.5mm}
    \label{det_3datasets}
    \setlength{\tabcolsep}{4.5pt}
\resizebox{0.5\linewidth}{!}{
\begin{tabular}{cccc}
\toprule
      & MNIST & Fashion-MNIST & CIFAR-10 \\ \hline
ARAE\cite{salehi2021arae}  & 0.975 & 0.936         & 0.602    \\
OCSVM\cite{chen2001one} & 0.960  & 0.928         & 0.586    \\
DSVDD\cite{ruff2018deep} & 0.948 & 0.928         & 0.648    \\
LSA\cite{abati2019latent}   & 0.975 & 0.922         & 0.641    \\
MKD\cite{salehi2021multiresolution}   & 0.987 & 0.945         & \textbf{0.872}    \\
Ours(R18)  & 0.985  & 0.939        & 0.577    \\
Ours(wR50)  & \textbf{0.990}  & \textbf{0.948}         & 0.663    \\
\bottomrule
\end{tabular}
}
\end{table}

\subsubsection{Anomaly Localization}

Compared with image-level anomaly detection, localizing subtle anomaly regions is more challenging and more practical in real-world scenarios.
We report the performance with   metrics  AUROC and AUPRO  for anomaly localization on the MVTecAD dataset.
As shown in Table. \ref{loc_auroc_mvtec}, the proposed method  outperforms the other state-of-the-art methods in the average of AUROC.
However, the AUROC indicator directly classifies pixels, which is more inclined to localization of large anomaly regions rather than small nes.
On the contrary, AUPRO takes into account anomaly regions of different sizes more fairly.  Table. \ref{loc_aupro_mvtec} shows the localization performance of different methods under the AUPRO metric.
we see that our proposed method     achieves a maximum improvement of 33\% as compared to the other methods.
It is worth noting that STPM \cite{wang2021student}, a multi-level feature matching teacher-student network  shows very competitive results in terms of  both AUROC and AUPRO metrics.
Note that our method is slightly inferior to SPTM in some texture categories, while performs better than SPTM in object categories such as the transistor. This observation further reflects the effectiveness of hierarchical feature differences in anomaly detection tasks. In addition, it also suggests that feature information extracted from pre-trained networks is more fully utilized by the autoencoder architecture than the teacher-student network.

 \begin{table*}[!h]
 \centering
\newcommand{\tabincell}[2]{\begin{tabular}{@{}#1@{}}#2\end{tabular}}

\caption{AUROC for Anomaly localization on MVTecAD dataset}
    \vspace {-2.5mm}
    \label{loc_auroc_mvtec}
    \setlength{\tabcolsep}{4.5pt}

\resizebox{0.75\linewidth}{!}{
\begin{tabular}{ccccccccc}
\toprule
Category   & AE$_{ssim}$\cite{bergmann2018improving} & AE$_{l2}$\cite{bergmann2018improving} & \tabincell{c}{Patch\\SVDD\cite{yi2020patch}} & SPADE\cite{cohen2020sub}          & STPM\cite{wang2021student}           & PaDiM\cite{defard2021padim}          & \tabincell{c}{Ours\\(R18)} & \tabincell{c}{Ours\\(wR50)}     \\ \hline
Carpet     & 0.87   & 0.59 & 0.926      & 0.975          & 0.988          & \textbf{0.991} & 0.989    & 0.989          \\
Grid       & 0.94   & 0.90 & 0.962      & 0.937          & \textbf{0.990} & 0.973          & 0.980    & 0.983          \\
Leather    & 0.78   & 0.75 & 0.974      & 0.976          & \textbf{0.993} & 0.992          & 0.990    & 0.991          \\
Tile       & 0.59   & 0.51 & 0.914      & 0.874          & \textbf{0.974} & 0.941          & 0.919    & 0.951          \\
Wood       & 0.73   & 0.73 & 0.908      & 0.885          & \textbf{0.972} & 0.949          & 0.953    & 0.962          \\
Bottle     & 0.93   & 0.86 & 0.981      & 0.984          & \textbf{0.988} & 0.983          & 0.974    & 0.978          \\
Cable      & 0.82   & 0.86 & 0.968      & 0.972          & 0.955          & 0.967          & 0.964    & \textbf{0.975} \\
Capsule    & 0.94   & 0.88 & 0.958      & \textbf{0.990} & 0.983          & 0.985          & 0.981    & 0.982          \\
Hazelnut   & 0.97   & 0.95 & 0.975      & \textbf{0.991} & 0.985          & 0.982          & 0.981    & 0.984          \\
Metal nut  & 0.89   & 0.86 & 0.980      & \textbf{0.981} & 0.976          & 0.972          & 0.957    & 0.972          \\
Pill       & 0.91   & 0.85 & 0.951      & 0.965          & \textbf{0.978} & 0.957          & 0.961    & 0.974          \\
Screw      & 0.96   & 0.96 & 0.957      & 0.989          & 0.983          & 0.985          & 0.984    & \textbf{0.990} \\
Toothbrush & 0.92   & 0.93 & 0.981      & 0.979          & \textbf{0.989} & 0.988          & 0.982    & 0.983          \\
Transistor & 0.90   & 0.86 & 0.970      & 0.941          & 0.825          & \textbf{0.975} & 0.901    & 0.957          \\
Zipper     & 0.88   & 0.77 & 0.951      & 0.965          & \textbf{0.985} & \textbf{0.985} & 0.977    & 0.977          \\ \hline
Mean       & 0.87   & 0.82 & 0.957      & 0.965          & 0.970          & 0.975          & 0.966    & \textbf{0.977} \\
\bottomrule
\end{tabular}
}
\end{table*}

\begin{table*}[!h]
\centering
\newcommand{\tabincell}[2]{\begin{tabular}{@{}#1@{}}#2\end{tabular}}
\caption{AUPRO \cite{bergmann2020uninformed} for Anomaly localization on MVTecAD dataset}
    \vspace {-2.5mm}
    \label{loc_aupro_mvtec}
    \setlength{\tabcolsep}{4.5pt}

\resizebox{0.75\linewidth}{!}{
\begin{tabular}{ccccccccc}
\toprule
Category   & AE$_{ssim}$\cite{bergmann2018improving} & AE$_{l2}$\cite{bergmann2018improving} & \tabincell{c}{Patch\\SVDD\cite{yi2020patch}} & SPADE\cite{cohen2020sub}          & STPM\cite{wang2021student}           & PaDiM\cite{defard2021padim}          & \tabincell{c}{Ours\\(R18)} & \tabincell{c}{Ours\\(wR50)}     \\ \hline
Carpet     & 0.647  & 0.456 & 0.695      & 0.947          & 0.958          & \textbf{0.962} & 0.957    & 0.952          \\
Grid       & 0.849  & 0.582 & 0.819      & 0.867          & \textbf{0.966} & 0.946          & 0.933    & 0.944          \\
Leather    & 0.561  & 0.819 & 0.819      & 0.972          & \textbf{0.980} & 0.978          & 0.976    & 0.976          \\
Tile       & 0.175  & 0.897 & 0.912      & 0.759          & \textbf{0.921} & 0.860          & 0.727    & 0.798          \\
Wood       & 0.605  & 0.727 & 0.725      & 0.874          & \textbf{0.936} & 0.911          & 0.910    & 0.908          \\
Bottle     & 0.834  & 0.910 & 0.918      & \textbf{0.955} & 0.951          & 0.948          & 0.914    & 0.921          \\
Cable      & 0.478  & 0.825 & 0.865      & 0.909          & 0.877          & 0.888          & 0.913    & \textbf{0.926} \\
Capsule    & 0.860  & 0.862 & 0.916      & \textbf{0.937} & 0.922          & 0.935          & 0.891    & 0.905          \\
Hazelnut   & 0.916  & 0.917 & 0.937      & 0.954          & 0.943          & 0.926          & 0.969    & \textbf{0.976} \\
Metal nut  & 0.603  & 0.830 & 0.895      & 0.944          & 0.945          & 0.856          & 0.984    & \textbf{0.994} \\
Pill       & 0.830  & 0.893 & 0.935      & 0.946          & \textbf{0.965} & 0.927          & 0.931    & 0.947          \\
Screw      & 0.887  & 0.754 & 0.928      & \textbf{0.960} & 0.930          & 0.944          & 0.928    & 0.950          \\
Toothbrush & 0.784  & 0.822 & 0.863      & \textbf{0.935} & 0.922          & 0.931          & 0.858    & 0.865          \\
Transistor & 0.725  & 0.728 & 0.701      & 0.874          & 0.695          & 0.845          & 0.773    & \textbf{0.885} \\
Zipper     & 0.665  & 0.839 & 0.933      & 0.926          & 0.952          & \textbf{0.959} & 0.934    & 0.931          \\ \hline
Mean       & 0.694  & 0.790 & 0.857      & 0.917          & 0.921          & 0.921          & 0.906    & \textbf{0.925} \\
\bottomrule
\end{tabular}
}
\end{table*}

\begin{figure}[!h]
\vspace {-4mm}
	\centering
		\includegraphics[width=4.9cm]{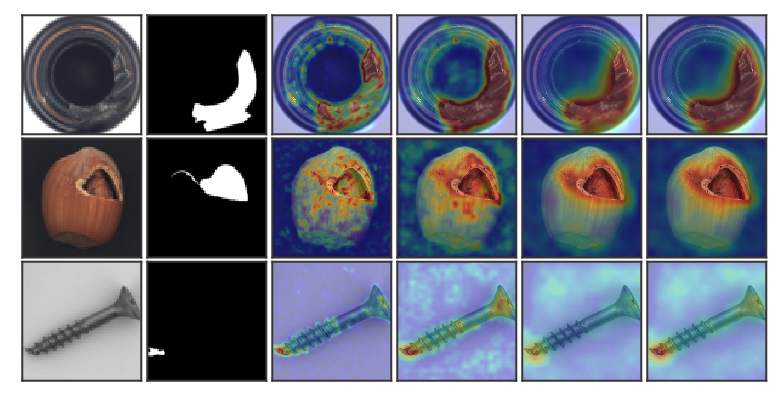}
	\caption{The qualitative result of the residual feature maps of our method on the MVTecAD dataset. Input images, ground truth regions, residual feature maps $\{ \phi_1, \phi_2, \phi_3\}$, and anomaly maps are displayed along the column direction.}
\label{FIG:3}
\end{figure}

\begin{figure}[!h]
	\centering
        \subfigure[Tile]{
		      \includegraphics[width=3.9cm]{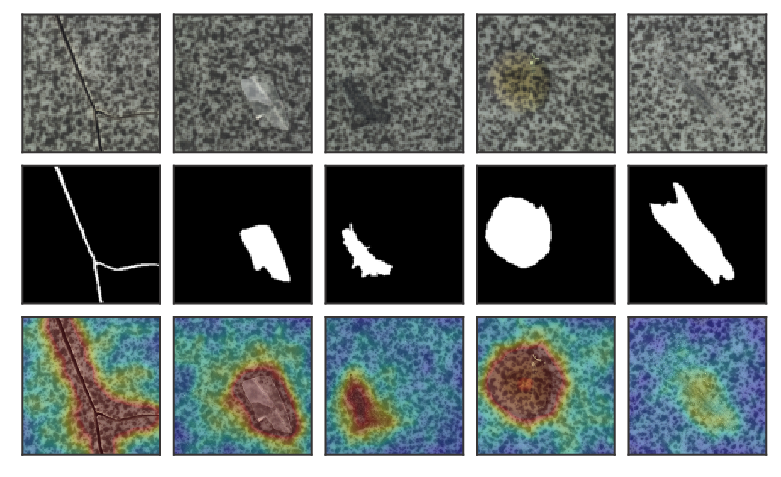}
        }
        \subfigure[Transistor]{
          \includegraphics[width=3.9cm]{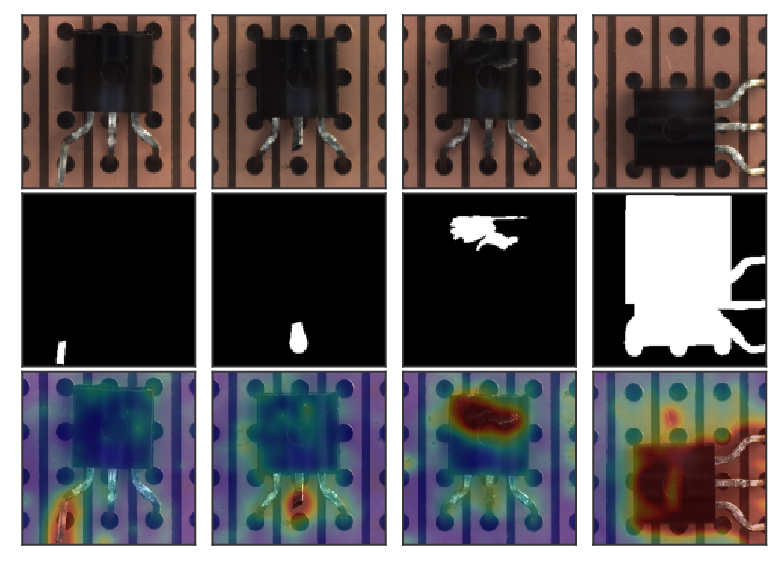}
        }
	\caption{Localization results of test examples with different types of anomalies, i.e. crack, glue strip, gray stroke, oil and rough in (a) tile and bent lead, cut lead, damaged case and misplaced case in (b) transistor.}
	\label{FIG:tile_transistor}
\end{figure}


Anomaly maps and residual feature maps $\{ \phi_1, \phi_2, \phi_3\}$ of proposed approach are visualized in Fig. \ref{FIG:3}. The residual map focuses on subtle anomalies with low-level features such as contours and edges of images, but is sensitive to noise. Instead, $\phi_3$ is relatively insensitive to noise with a smoother value variation and has a large receptive field with strong semantics. Fig. \ref{FIG:tile_transistor} shows how our approach localizes anomalous images of tile and transistor with different types of anomalies. For instance, detecting tiny cracks in tiles requires spatially rich information from the encoder, while misplaced transistors require strong semantic information. The combination of residual maps at multiple resolutions enhances performance for most categories of datasets. However, our method may generate some anomaly scores that affect performance in images with regions with complex features similar to noise.

\section{Conclusions}
We propose a simple yet effective method using an encoder-decoder architecture to reconstruct the hierarchical feature representations of the input image, and then compute residual maps between the encoded features and the reconstructed features at different layers to detect and locate anomalies. Without any extra labels and data augmentation in the training phase, our method outperforms the  state-of-the-art methods on MNIST, Fashion-MNIST, CIFAR-10, and MVTecAD dataset.

\bibliographystyle{splncs04}
\bibliography{refer}
%





\end{document}